\documentclass[sigconf]{acmart}
\usepackage{graphicx}
\usepackage{caption}
\usepackage{xcolor}
\usepackage{enumitem}
\usepackage{fontawesome5}
\usepackage[most]{tcolorbox}
\usepackage{xspace}

\definecolor{headerblue}{RGB}{0, 102, 204}
\definecolor{blockgray}{RGB}{245, 245, 245}

\newtcolorbox{convbox}{colback=blockgray, colframe=headerblue, title style={color=white, fill=headerblue}, boxrule=0.5pt, fonttitle=\bfseries, left=5pt, right=5pt}

\begin{document}

\title{MIRA: A Novel Framework for Fusing Modalities in Medical RAG}

\author{Jinhong Wang}
\authornote{Both authors contributed equally to this research.} 
\affiliation{%
  \institution{Department of Computer Vision, MBZUAI}
  \city{Abu Dhabi}
  \country{United Arab Emirates}
}
\email{jinhong.wang@mbzuai.ac.ae}

\author{Tajamul Ashraf}
\authornote{Corresponding author.}
\authornotemark[1] 
\affiliation{%
  \institution{Department of Computer Vision, MBZUAI}
  \city{Abu Dhabi}
  \country{United Arab Emirates}
}
\email{tajamul.ashraf@mbzuai.ac.ae}

\author{Zongyan Han}
\affiliation{%
  \institution{Department of Computer Vision, MBZUAI}
  \city{Abu Dhabi}
  \country{United Arab Emirates}
}
\email{zongyan.han@mbzuai.ac.ae} 

\author{Jorma Laaksonen}
\affiliation{%
  \institution{Department of Computer Science, Aalto University}
  \city{Aalto}
   \country{Finland}
}
\email{jorma.laaksonen@aalto.fi} 

\author{Rao Mohammad Anwer}
\affiliation{%
  \institution{Department of Computer Vision, MBZUAI}
  \city{Abu Dhabi}
  \country{United Arab Emirates}
}
\email{rao.anwer@mbzuai.ac.ae} 



\def\mllm{\texttt{MLLM}\xspace}
\def\rag{\texttt{RAG}\xspace}
\def\llm{\texttt{LLM}\xspace}
\def\ai{\texttt{AI}\xspace}
\def\vqa{\texttt{VQA}\xspace}
\def\medvqa{\texttt{MedVQA}\xspace}

\begin{abstract}


Multimodal Large Language Models (\mllm) have significantly advanced AI-assisted medical diagnosis but often generate factually inconsistent responses that deviate from established medical knowledge. Retrieval-Augmented Generation (\rag) enhances factual accuracy by integrating external sources, but it presents two key challenges. First, insufficient retrieval can miss critical information, whereas excessive retrieval can introduce irrelevant or misleading content, disrupting model output. Second, even when the model initially provides correct answers, over-reliance on retrieved data can lead to factual errors. To address these issues, we introduce Multimodal Intelligent Retrieval and Augmentation (\textbf{MIRA}) framework designed to optimize factual accuracy in \mllm. \textbf{MIRA} consists of two key components: (1) a calibrated Rethinking and Rearrangement module that dynamically adjusts the number of retrieved contexts to manage factual risk, and (2) A medical \rag  framework integrating image embeddings and a medical knowledge base with a query-rewrite module for efficient multimodal reasoning. This enables the model to effectively integrate both its inherent knowledge and external references. Our evaluation of publicly available medical VQA and report generation benchmarks demonstrates that \textbf{MIRA} substantially enhances factual accuracy and overall performance, achieving new state-of-the-art results. Code, is released at \href{https://github.com/mbzuai-oryx/MIRA}{\textcolor{blue}{https://github.com/mbzuai-oryx/MIRA}}.


\end{abstract}   


\maketitle

\section{Introduction}
\label{sec:intro}
\begin{figure}[t]
    \centering
    \includegraphics[width=\linewidth]{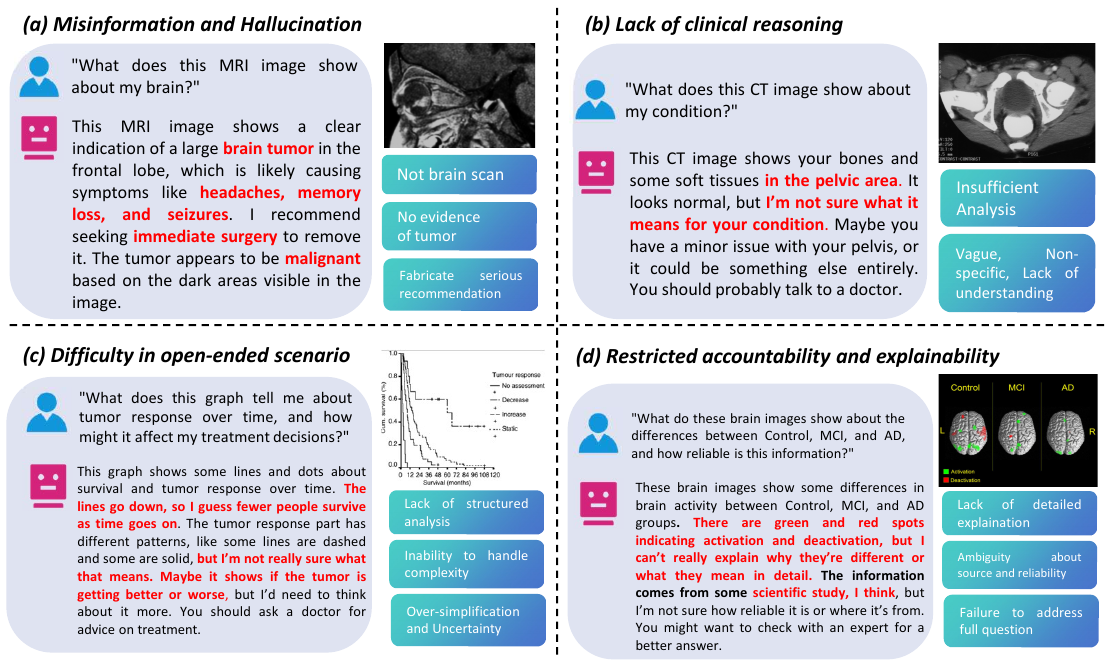} \vspace{-0.5cm}
    \caption{Overview of key constraints in automatic medical question answering, highlighting the challenges related to factuality, domain specificity, and clinical relevance that must be addressed for reliable system performance.} \vspace{-0.3cm}
    \label{fig:teasor1}
\end{figure}

Medical Visual Question Answering (\texttt{MedVQA}) plays a crucial role in AI-driven healthcare by enabling accurate diagnosis, disease detection, treatment optimization, and clinical decision-making~\cite{vqamed, ben2019vqa, liu2023q2atransformer, ben2021overview}. The integration of multimodal data, such as medical images and corresponding textual reports, presents unique challenges like domain shifts~\cite{020303}, limited labeled data~\cite{gu2021domain, ashraf2025fatefocalmodulatedattentionencoder}, and modalities~\cite{lin2022medical}, etc, which require a balance between visual reasoning and textual coherence. Existing models, including \textit{LLaVA-Med}~\cite{li2024llava} and \texttt{Huatuo-Vision-7B}~\cite{wang2023huatuo}, struggle with hallucinations, rigid reasoning, and unimodal biases, often misinterpreting subtle imaging findings and failing in open-ended queries.   Although Multimodal Large Language Models (\texttt{MLLMs})~\cite{MLLMinHealthCare, demirhan2023survey, moor2023med} have shown resilience to internal knowledge that often leads to \textit{hallucinations} and factual inaccuracies that pose risks to patient safety \cite{park2024patient}. Representative analyses of these constraints are shown in Fig.~\ref{fig:teasor1}, and these issues compromise their reliability in medical use. And there's calls for more reliable and precise models to perform \texttt{MedVQA}. To address this issue, Retrieval-Augmented Generation (\texttt{RAG})~\cite{xia2024mmedrag, 020401} has emerged as a promising solution by enriching model responses with verified external medical literature \cite{xia2024rule}. This approach enhances factual grounding, reduces hallucinations, and improves the reliability of AI-driven clinical insights~\cite{xia2024mmedrag}.  

Despite its potential, \texttt{RAG} introduces two critical challenges in medical settings: (1) \textbf{Retrieval imbalance}, where the system disproportionately relies on specific sources, potentially overlooking relevant yet underrepresented data, leading to biased or incomplete information synthesis; and (2) \textbf{Over-reliance}, where the model excessively trusts the retrieved information without adequate verification, increasing the risk of propagating factual inaccuracies and medical misinformation.  

Recently, various methods~\cite{xia2024rule, xia2024mmedrag, xraygpt} have attempted to address these issues by incorporating retrieval calibration and preference tuning to mitigate over-reliance. While these strategies improve factual accuracy to some extent, they have notable limitations. Specifically, prior approaches are predominantly designed for radiology-related tasks and often struggle to generalize across diverse medical domains such as pathology, dermatology, and cross-specialty diagnostic reasoning as shown in Fig.~\ref{fig:teasor1}. Prior approaches~\cite{xia2024mmedrag, xia2024rule} decouple retrieval from reasoning, leading to fragmented outputs without dynamic modality fusion or real-time evidence integration. Furthermore, these frameworks typically treat vision and text as unified modalities, neglecting the need for modality-specific processing and adaptive fusion. More critically, they rely on static databases for retrieval, lacking mechanisms for real-time evidence integration, which is essential for ensuring up-to-date medical knowledge. These limitations hinder their effectiveness in complex clinical scenarios requiring dynamic, multimodal reasoning.

To address the challenges in multimodal medical reasoning, we introduce Multimodal Intelligent Retrieval and Augmentation (\textbf{MIRA}), a novel \texttt{RAG} framework that enhances clinical decision-making through structured retrieval, validation, and reasoning. Our approach integrates two key innovations: (1) an \textbf{RTRA pipeline} that employs iterative ``rethink-rearrange'' cycles to refine outputs by validating them against retrieved multimodal medical knowledge, ensuring factual accuracy and interpretability, and (2) a \textbf{Multimodal RAG (MRAG)} strategy that combines \textit{offline} (prebuilt medical databases) and \textit{online} (real-time web sources) retrieval to ground responses in dynamically updated clinical evidence. The MRAG pipeline jointly retrieves textual and visual information from sources like NIH ChestX-ray14 and real-time medical guidelines, while the \texttt{RTRA} cycle ensures clinical consistency through a three-stage validation process: (i) initial response generation, (ii) self-critique via Chain-of-Thought (CoT), and (iii) final answer refinement. Our vision-language backbone, built on \texttt{CLIP-ViT-Large} and \texttt{SigLIP}, is trained on 500K medical image-text pairs, enabling robust multimodal alignment. Our contributions include a novel Vision-Language Model (\texttt{VLM}) extending \texttt{LLaVA} with dynamic modality fusion, the \textbf{first multimodal RAG framework} for \texttt{MedVQA}, and state-of-the-art performance across multiple benchmarks, achieving \textit{9 times faster inference} than 72B models, making it efficient for real-world clinical applications.



Our main contributions are as follows:
\begin{itemize}
    \item  We introduce \textbf{MIRA}, the first retrieval-augmented generation (RAG) framework that seamlessly integrates structured multimodal retrieval with \textit{adaptive reasoning}, surpassing existing methods in clinical decision-making accuracy and efficiency.
    \item  Unlike static retrieval paradigms, our \textit{context-rethink module} employs an iterative ``rethink-rearrange'' cycle for \textit{dynamic $k$-selection}, ensuring precision in evidence selection, while \textit{Chain-of-Thought (CoT)} reasoning enhances factual consistency in MedVQA.
    \item  Our architecture pioneers a dual-pathway retrieval mechanism with specialized vision and language encoders, enabling \textit{fine-grained image-text alignment}. The integration of a \textit{curated citation module} ensures interpretability, setting a new standard for transparency in medical AI.
    \item  Our model establishes new state-of-the-art (SOTA) results on MedVQA, significantly reducing factual errors through \textit{online search augmentation} and \textit{adaptive CoT-based verification}. With \textbf{9× faster inference} than 72B models, it sets a new benchmark for real-time, high-precision medical reasoning.
    
\end{itemize}

\section{Related Work}
\label{sec:related_work}

\noindent\textbf{Multimodal Large Language Models (MLLMs):}
Traditional LLMs are limited to textual input, restricting their ability to process and understand multimodal information. Multimodal LLMs extend their capabilities by incorporating additional modalities such as images, videos, and audio, enabling a richer and more holistic understanding of input data~\cite{Thirunavukarasu2023LargeLM, zhang2023biomedgpt, MLLMinHealthCare, peng2023study}. 
A typical MLLM architecture consists of three primary components: a modality encoder, a pre-trained LLM, and a modality generator, interconnected through learned projectors to facilitate seamless information transformation across different modalities~\cite{demirhan2023survey, moor2023med}. These architectures have been successfully applied in various domains, including healthcare, where multimodal integration has enhanced clinical decision support, diagnosis, and medical research applications.

\noindent\textbf{Medical Large Vision-Language Models (Med-LVLMs):}
Medical Large Vision-Language Models (Med-LVLMs) are specialized architectures designed to process and interpret medical images alongside textual data. These models typically integrate an LLM with a dedicated vision module that extracts relevant information from medical images and converts it into a representation compatible with the LLM’s processing capabilities. 
Given a medical image $x_v$ and a clinical query $x_t$, the combined input is represented as $x = (x_v, x_t)$. The model then autoregressively predicts the probability distribution of the text output $y$ based on the multimodal input. Med-LVLMs have been applied in tasks such as radiology report generation, pathology image analysis, and multi-modal medical question answering~\cite{zhang2023biomedclip, peng2023study, moor2023med}. Unlike these, MIRA integrates iterative reasoning (RTRA) and multimodal retrieval-augmented generation (MRAG) to ensure that outputs are validated against retrieved evidence, addressing the issue of hallucination and improving accountability. By introducing a dynamic multimodal RAG pipeline that adapts to real-time evidence, MIRA enhances both the versatility and precision of medical vision-language models.


\noindent\textbf{Retrieval-Augmented Generation (RAG):}
Retrieval-Augmented Generation (RAG) has emerged as an effective paradigm to address LLMs' limitations in domain-specific question answering~\cite{xia2024mmedrag, xraygpt, xia2024rule, 020402}. By combining an LLM with an external knowledge retriever, RAG allows models to access relevant information dynamically, improving accuracy and factual consistency in generated responses. This approach has been successfully applied to tasks such as question answering, fact verification, and knowledge-intensive text generation, achieving state-of-the-art performance in open-domain scenarios~\cite{vqamed, ben2019vqa, liu2023q2atransformer}.
Recent advances in multimodal RAG have further expanded its capabilities by incorporating images, videos, and structured data into retrieval mechanisms~\cite{chen2022muragmultimodalretrievalaugmentedgenerator, lin2023pmcclip, karpukhin2020dense}. Multimodal RAG frameworks have been explored in medical applications, where the integration of medical text with radiological and histopathological images has improved diagnostic accuracy and clinical decision support~\cite{xia2024rule, lin2022medical, 010202}. While existing works like MMed-RAG~\cite{xia2024mmedrag} and RULE~\cite{xia2024rule} address factuality challenges, they rely on static retrieval mechanisms. MIRA extends this paradigm by incorporating online retrieval and iterative reasoning, ensuring that outputs are dynamically refined against cross-modal evidence, thereby enhancing adaptability and reliability in real-world clinical scenarios.

\noindent\textbf{Evaluation of Retrieval-Augmented Generation (RAG):}
Evaluating RAG models involves evaluating both the retrieval and the generation components. Frameworks like RAGAs~\cite{es2023ragasautomatedevaluationretrieval} propose metrics such as Faithfulness, Answer Relevance, and Context Relevance to provide a comprehensive evaluation. Benchmarking datasets such as TruthfulQA~\cite{010203} and MMLU~\cite{devlin2019bert} have been widely used to measure retrieval performance and factual accuracy. Although automated evaluation methods offer scalability and alignment with human judgments, they remain approximations. Human annotations remain the gold standard, particularly in domain-specific applications like medical question answering, where accuracy is critical. For multimodal evaluation, GPT-4V has shown a strong alignment with human assessments in vision language tasks~\cite{zhang2023biomedgpt}, further highlighting the potential of large multimodal models in knowledge-intensive applications. Despite these advancements, existing RAG-based evaluation frameworks often struggle with dynamic multimodal reasoning and ensuring real-time validation of retrieved knowledge. MIRA addresses this limitation by integrating iterative reasoning with online retrieval, allowing the model to refine outputs against cross-modal evidence continuously. This not only improves factual accuracy and faithfulness but also enhances the adaptability of the system to real-world clinical settings. By leveraging retrieval-aware generation and multimodal verification, our approach sets a new benchmark in robust, evidence-grounded medical AI.

%
\section{Methodology}
\label{sec:method}

\begin{figure*}[t]
    \centering
    \includegraphics[width=\linewidth]{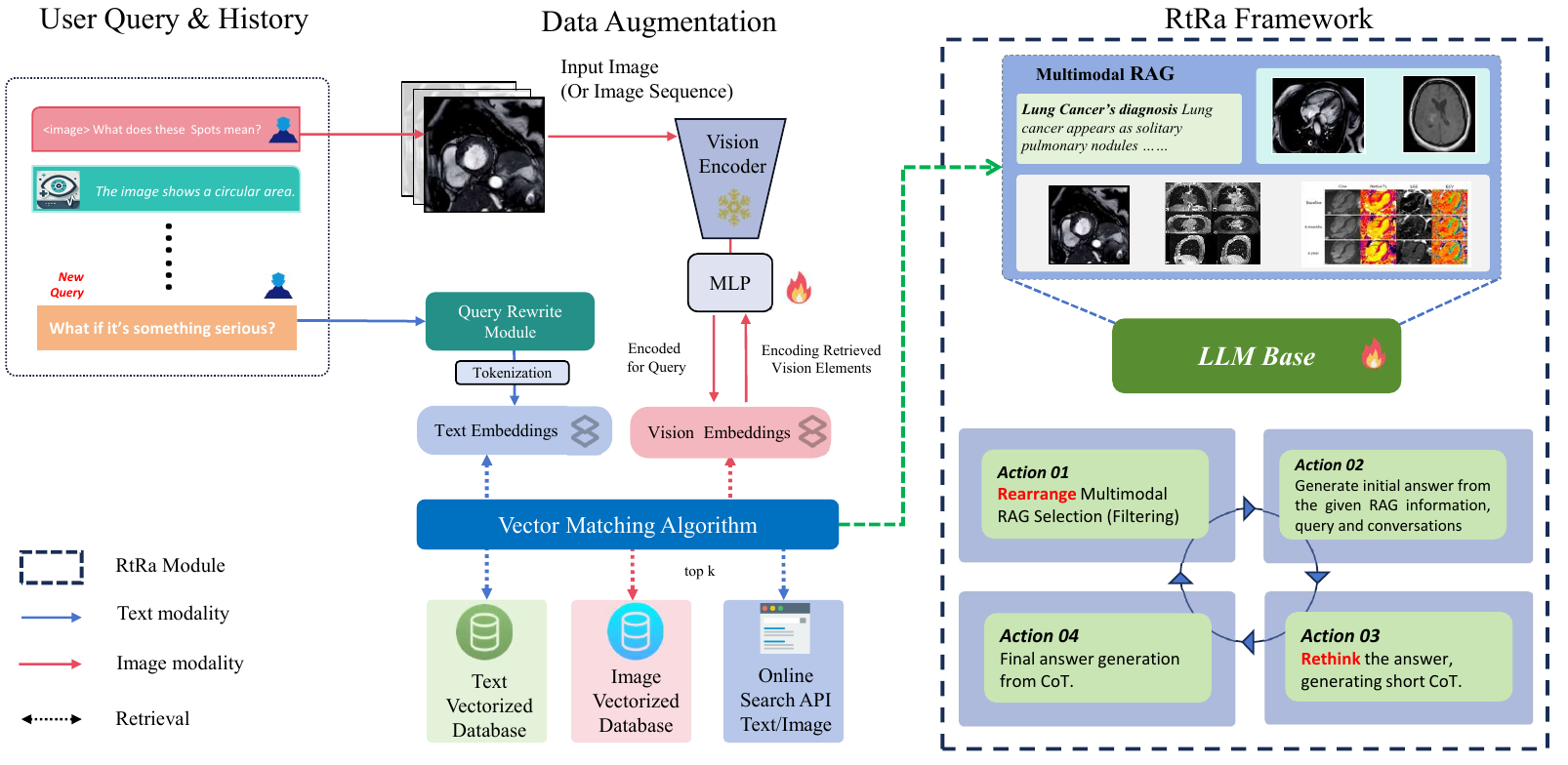}
    \caption{Overview of the MIRA (Multimodal Intelligent Retrieval and Augmentation) pipeline. The system integrates image and text-based retrieval to enhance the generation process.}
    \label{fig:multimodal_rag}
\end{figure*}

In this section, we introduce MIRA, a novel framework designed to enhance multimodal reasoning in medical applications, particularly MedVQA. Unlike conventional RAG systems, MIRA optimizes the retrieval, fusion, and generation of both textual and visual data to improve factual accuracy and interoperability as illustrated in Fig.~\ref{fig:multimodal_rag}. Built upon the LLaVA architecture, MIRA integrates a high-resolution vision encoder for detailed medical image understanding and introduces dynamic modality fusion to adaptively balance textual and visual contributions based on the query type. Furthermore, it employs a multimodal RAG pipeline, retrieving both text and images from structured and unstructured medical databases using a joint embedding space. Finally, MIRA enhances reasoning through a Rethink-Rearrange (RTRA) mechanism, leveraging Chain-of-Thought (CoT) reasoning for iterative refinement, ensuring responses are clinically accurate and well-supported by retrieved evidence. We will now discuss each component of MIRA in detail.

\subsection{Data preprocessing}

\textbf{Multimodal Input. }The MIRA framework receives two primary inputs: a textual query \( Q \) and an image \( I \). These inputs are processed through separate pathways—textual and visual—and then combined in a manner that allows for sophisticated multimodal reasoning and retrieval-augmented generation (RAG). The inputs are transformed into high-dimensional embeddings through the use of a text and vision encoder. The embeddings serve as the basis for data retrieval and fusion during the reasoning process.

\noindent\textbf{Query Rewrite Module. } Before the input query \( Q \) is passed to the text encoder, it undergoes an initial transformation via the Query Rewrite module. This module aims to enhance the semantic alignment and contextual clarity of the input query. Specifically, given an input query \( Q \), the module applies a learned function \( \mathcal{R}(\cdot) \) to rewrite and refine the query \( Q \), resulting in a modified query \( Q' \). The rewriting function can be expressed as:
\[
Q' = \mathcal{R}(Q), \quad \mathcal{R} : \mathbb{R}^{n_{\text{words}}} \rightarrow \mathbb{R}^{n'_{\text{words}}},
\]
where \( \mathcal{R}(\cdot) \) is a transformation that adjusts the query’s linguistic structure while preserving its semantic meaning. The rewritten query \( Q' \) serves as an enriched representation, ensuring it is aligned with the task requirements.

\noindent\textbf{Text Encoder. }Following the query refinement in the Query Rewrite module, the enhanced query \( Q' \) is passed to the text encoder \( \mathcal{E}_{\text{text}} \). This encoder operates as a learned function that maps the refined query \( Q' \) to a high-dimensional semantic embedding \( E_{\text{text}}' \in \mathbb{R}^{d_{\text{text}}} \), where \( d_{\text{text}} \) is the dimensionality of the embedding space. This process effectively captures the deeper semantic structure and contextual information of the query, facilitating its integration into the multimodal reasoning process.

The text encoder can be formulated as a transformation function \( \mathcal{E}_{\text{text}} : \mathbb{R}^{n'_{\text{words}}} \rightarrow \mathbb{R}^{d_{\text{text}}} \), which takes the rewritten query \( Q' \in \mathbb{R}^{n'_{\text{words}}} \) as input and produces the embedding \( E_{\text{text}}' \in \mathbb{R}^{d_{\text{text}}} \):
\[
E_{\text{text}}' = \mathcal{E}_{\text{text}}(Q'), \quad E_{\text{text}}' \in \mathbb{R}^{d_{\text{text}}}.
\]
Here, \( n'_{\text{words}} \) denotes the number of words in the rewritten query \( Q' \), and \( d_{\text{text}} \) represents the dimensionality of the learned embedding space. The encoder \( \mathcal{E}_{\text{text}} \) is a pre-trained transformer model  optimized to capture both syntactic and semantic features of the query.
The resulting text embedding \( E_{\text{text}}' \) serves as a rich, high-dimensional representation of the query, capturing its meaning in a manner that is agnostic to the specific wording of the query. This embedding is essential for understanding the query in the context of a multimodal task, where it will be fused with corresponding visual information to perform retrieval or reasoning tasks in the MIRA framework.

\noindent\textbf{Vision Encoder. } Parallel to the textual processing pipeline, the input image \( I \in \mathbb{R}^{H \times W \times 3} \) is passed through the vision encoder \( \mathcal{E}_{\text{vision}} \), which is designed to extract rich visual features from the raw image. The encoder \( \mathcal{E}_{\text{vision}} \) maps the input image to a high-dimensional feature space \( E_{\text{image}}' \in \mathbb{R}^{d_{\text{image}}'} \), where \( d_{\text{image}}' \) is the dimensionality of the image embedding space. This transformation is mathematically defined as:
\[
E_{\text{image}}' = \mathcal{E}_{\text{vision}}(I), \quad E_{\text{image}}' \in \mathbb{R}^{d_{\text{image}}'}.
\]
The vision encoder \( \mathcal{E}_{\text{vision}} \) utilizes the CLIP-ViT-Large model~\cite{clip}, which is based on the Vision Transformer (ViT) framework. The model is optimized to capture complex spatial patterns, textures, and object relationships within the image, which are essential for tasks that require vision and language understanding structured as follows:
\[
E_{\text{image}}' = f_{\text{ViT}}(I), \quad f_{\text{ViT}}: \mathbb{R}^{H \times W \times 3} \rightarrow \mathbb{R}^{d_{\text{image}}'},
\]
where \( f_{\text{ViT}} \) denotes the Vision Transformer-based function applied to the input image \( I \). The resulting image embedding \( E_{\text{image}}' \) serves as a high-dimensional representation of the image content, encoding important visual information such as object features, textures, and spatial relationships.

MIRA further enhances the LLaVA framework by improving the precision of the vision encoder. The CLIP-ViT-Large model processes high-resolution images \( I' \in \mathbb{R}^{H' \times W' \times 3} \), where \( H' > H \) and \( W' > W \), allowing the model to capture finer visual details. The enhanced vision encoder now generates a refined image embedding \( E'_{\text{image}} \in \mathbb{R}^{d_{\text{image}}'} \), which improves the model's ability to handle complex visual reasoning tasks. Mathematically, the operation is represented as:
\[
E'_{\text{image}} = \mathcal{E}_{\text{vision}}(I'), \quad E'_{\text{image}} \in \mathbb{R}^{d_{\text{image}}'}.
\]
Thus, the upgraded vision encoder with higher resolution images results in a more expressive visual feature space, better suited for challenging multimodal tasks that require detailed image understanding.

\subsection{Multimodal Database}

\noindent\textbf{Data Retrieval. }
Once the textual and visual embeddings \( E_{\text{text}}' \) and \( E_{\text{image}}' \) are computed, MIRA employs a data retrieval mechanism to search for the most relevant multimodal entries from both offline and online databases. The two primary sources of data retrieval are:
1. \textbf{Offline Database:} A collection of pre-indexed text and image embeddings, denoted as \( \mathcal{D}_{\text{offline}} = \{(T_{\text{offline},i}, I_{\text{offline},i})\} \), where each pair \( (T_{\text{offline},i}, I_{\text{offline},i}) \) represents a text and its associated image in the offline database.
2. \textbf{Online API Database:} A set of real-time data entries retrieved via an API, denoted as \( \mathcal{D}_{\text{API}} = \{(T_{\text{API},i}, I_{\text{API},i})\} \), where each pair \( (T_{\text{API},i}, I_{\text{API},i}) \) is a dynamically fetched text-image pair.

The retrieval function \( \text{Retrieve}(E_{\text{text}}', E_{\text{image}}') \) computes the similarity between the query embeddings \( E_{\text{text}}' \) and \( E_{\text{image}}' \), and the embeddings stored in both the offline and online databases. This is achieved using vector matching algorithms, such as cosine similarity, to measure the closeness between the query embeddings and the stored entries in the respective databases:
\[
e_{\text{retrieved},j} = \text{Retrieve}(E_{\text{text}}', E_{\text{image}}', \mathcal{D}_{\text{offline}}, \mathcal{D}_{\text{API}}),
\]
where \( e_{\text{retrieved},j} \in \mathcal{D}_{\text{offline}} \cup \mathcal{D}_{\text{API}} \) represents the \( j \)-th retrieved entry from either database.

Next, the top \( k \) relevant entries, consisting of both text and image pairs, are retrieved. The number \( k \) represents the number of top entries to retrieve, where typically the retrieval involves:
- \( k_{\text{text}} = 3 \) retrieved text chunks, denoted as \( T_{\text{retr},1}, T_{\text{retr},2}, \dots, T_{\text{retr},k_{\text{text}}} \),
- \( k_{\text{image}} = 2 \) retrieved images, denoted as \( I_{\text{retr},1}, I_{\text{retr},2} \).

In addition to these, the API also retrieves a dynamically fetched text paragraph \( T_{\text{API\_retrieved}} \), which provides an additional source of textual data for the fusion stage.
Once the relevant entries are retrieved, the fusion process integrates the textual and visual embeddings from both the query and the retrieved database entries. The fusion of the query and retrieved embeddings is performed using a dynamic attention mechanism, which adaptively assigns different weights to the contributions of the text and image embeddings based on their relevance to the query.

\noindent\textbf{Fusion of Retrieved Embeddings.} After retrieving the top-k relevant entries from the multimodal database, MIRA employs a dynamic attention mechanism to fuse the image and text embeddings. The attention mechanism adapts to the nature of the query, adjusting the relative contributions of each modality to ensure the most pertinent features are prioritized.

The final fused representation \( E_{\text{final}} \) is computed as:
\[
E_{\text{final}} = \text{Att}(E_{\text{image}}, E_{\text{text}}), \quad E_{\text{final}} \in \mathbb{R}^{d_{\text{final}}},
\]
where \( \text{Att()} \) represents the attention mechanism, which computes the weighted combination of the visual (\( E_{\text{image}} \)) and textual (\( E_{\text{text}} \)) embeddings. The resulting \( E_{\text{final}} \) is a multimodal representation, where the contributions of text and image are adaptively balanced based on the query context.
This dynamic fusion enables MIRA to prioritize either modality based on the specific nature of the query. For example, if the query is more text-centric, the model gives higher weight to the textual representation, while for image-centric queries, the visual features are emphasized.

\noindent\textbf{Alignment layer and Modality Fusion. }Once the image and text embeddings are computed, MIRA employs an adaptive attention mechanism to align these modalities into a shared multimodal space. Unlike other frameworks, such as LLaVA, which utilize simple linear transformations for alignment, MIRA’s attention-based fusion mechanism allows for dynamic adjustment of the weight given to each modality. To align the image and text representations in a common multimodal space, MIRA computes the final multimodal embedding \( E_{\text{final}} \) as a weighted combination of the image and text embeddings. This is formalized as:
\[
E_{\text{final}} = \text{Att}(E_{\text{image}}, E_{\text{text}}), \quad E_{\text{final}} \in \mathbb{R}^{d_{\text{final}}},
\]
where \( \text{Att}(\cdot) \) is the attention function that computes the final fused representation by attending to both the image and text features. The fusion ensures that the multimodal representation captures the relevant features from both modalities, which are contextually weighted based on the query.

In image-centric queries (e.g., "Describe this chest X-ray"), the attention mechanism increases the weight on the image representation \( E_{\text{image}} \), while in text-centric queries (e.g., "What is the symptom of this disease?"), the model prioritizes the text representation \( E_{\text{text}} \).

\noindent\textbf{Modality-Specific Query Processing. }
Given a query \( Q \) and an associated image \( I \), the model determines the most relevant modality to focus on based on the nature of the query. This is achieved through an attention function that dynamically selects the contribution from each modality. The fusion of the image and text embeddings is adjusted using a learnable parameter \( \alpha \), which defines the balance between the two modalities.

The final multimodal representation is computed as:
\[
E_{\text{final}} = \alpha \cdot \text{Att}(E_{\text{image}}) + (1 - \alpha) \cdot \text{Att}(E_{\text{text}}),
\]
where \( \alpha \in [0, 1] \) is a learnable parameter that determines the weighting of the image and text features. When \( \alpha \) is close to 1, the model focuses more on the image, while if \( \alpha \) is close to 0, the model prioritizes the text features.  Finally, the processed input now consists of the following multimodal entries:
\begin{table}[h]
    \centering
    \renewcommand{\arraystretch}{1.0} 
    \setlength{\tabcolsep}{8pt} 
    \resizebox{\columnwidth}{!}{ 
    \begin{tabular}{ll}
        \toprule
        \textbf{Type} & \textbf{Description} \\
        \midrule
        \( I_{\text{original}} \) & Original Image \\
        \( T_{\text{original}} \) & Original Text \\
        \( \{T_{\text{retrieved},1}, T_{\text{retrieved},2}, T_{\text{retrieved},3}\} \) & Retrieved Text Chunks \\
        \( \{I_{\text{retrieved},1}, I_{\text{retrieved},2}\} \) & Retrieved Images \\
        \( T_{\text{API\_retrieved}} \) & API Retrieved Text Paragraph \\
        \bottomrule
    \end{tabular}
    }
    \caption{Structured Input Components for Multimodal RAG}
    \label{tab:input_components}
    
\end{table}
This combined set forms the new user input for the subsequent reasoning steps in the MIRA framework, as shown in Table~\ref{tab:input_components}.

\subsection{Multimodal Medical RAG}

\noindent\textbf{RAG Pipeline. }The Medical MRAG Module is designed to process diverse medical data, integrating multimodal information to enhance diagnostic accuracy and provide contextually relevant insights. The core of the retrieval mechanism is powered by a high-performance vector matching algorithm utilizing the Faiss-CUDA framework. The knowledge base $\mathcal{K}$ consists of vectorized representations of both text and images, enabling efficient similarity search. Given a query $q$, the system retrieves a relevant set of documents $\mathcal{D}$ from $\mathcal{K}$, represented as:

\begin{equation}
    \mathcal{D} = \{d_1, d_2, \dots, d_n\}, \quad d_i \in \mathcal{K}.
\end{equation}

Retrieval-Augmented Generation (RAG) plays a central role in refining the generative process. Once the relevant documents are retrieved, they are conditioned into the generation model to ensure domain-specific accuracy. The probability distribution of the generated medical response $y$ is computed as:

\begin{equation}
    p(y | q, \mathcal{D}) = \prod_{t=1}^{T} p(y_t | y_{<t}, q, \mathcal{D}; \theta),
\end{equation}

where $\theta$ represents the model parameters. The retrieved multimodal RAG information is formatted using the Llava joint embedding framework, after which the sequence undergoes further processing through a Multilayer Perceptron (MLP) to refine its structure and contextual coherence.

To maintain real-time applicability and continually expand the RAG database, the pipeline incorporates an Online Search API. This API leverages the DuckDuckGo open API to retrieve additional text and image data, ensuring that the system remains updated with the latest medical research and clinical guidelines. By integrating both static vectorized databases and dynamically retrieved online resources, the Medical MRAG Module achieves an optimal balance between efficiency and accuracy, making it a robust tool for analyzing complex medical imaging data and supporting diagnostic decision-making.

\noindent\textbf{Reinforced CoT Generation. } To enhance Chain-of-Thought (CoT) reasoning in medical question-answering, we introduce a reinforcement-driven multimodal retrieval-augmented generation (RAG) pipeline. The process begins with structured data selection across both offline and online RAG sources, ensuring optimal alignment between textual and visual modalities. Given input queries $q$—comprising text $q_t$ and image $q_i$—RAG data $\mathcal{D}$ is refined through a filtering function:

\begin{equation}
    \mathcal{D} = \{ d_i \in \mathcal{K} \mid \text{Rearrange}(d_i, q) \neq \texttt{<None>} \}.
\end{equation}

Here, \texttt{Rearrange} truncates extraneous data while ensuring at least one textual and visual reference remains relevant. The extracted $\mathcal{D}$ forms the basis for initial CoT generation.

Leveraging Qwen2.5-VL models, vision-text embeddings $f_v(q_i)$ and $f_t(q_t)$ are jointly processed, refining knowledge selection through multimodal alignment:

\begin{equation}
    \hat{\mathcal{D}} = \operatorname{argmax}_{d \in \mathcal{D}} \langle f_v(q_i) + f_t(q_t), f(d) \rangle.
\end{equation}

Following knowledge retrieval, the initial CoT response $y_0$ is generated:

\begin{equation}
    y_0 = p(y | q, \hat{\mathcal{D}}).
\end{equation}

To ensure coherence and factual correctness, an iterative refinement stage \texttt{Rethink} is applied, where an expert model reassesses $y_0$ against the retrieved knowledge $\hat{\mathcal{D}}$, producing an optimized response $y^*$:

\begin{equation}
    y^* = \operatorname{Refine}(y_0, \hat{\mathcal{D}}, \mathcal{K}).
\end{equation}

The entire process is reinforced using a reward function $R(z, y)$, which evaluates logical consistency and medical accuracy:

\begin{equation}
    R(z, y) = \lambda_1 \cdot \text{FactualScore}(z, \mathcal{K}) + \lambda_2 \cdot \text{CoherenceScore}(y, z),
\end{equation}

where $\lambda_1, \lambda_2$ control the balance between factual precision and coherence. Model parameters $\theta$ are updated via policy gradient optimization:

\begin{equation}
    \nabla_\theta J(\theta) = \mathbb{E}_{z \sim p_\theta} \left[R(z, y) \nabla_\theta \log p_\theta(z | q) \right].
\end{equation}

This reinforcement learning (RL)-guided CoT refinement ensures structured, high-quality medical reasoning. The training dataset is derived from LLaVA-Med’s 500k alignment set, with an additional 50k high-quality CoT fine-tuning instances curated from PubMedVision and VQA-Med.

\subsection{Optimization}

\section{Optimization with Reinforcement Learning and Chain-of-Thought Reasoning}

We optimize our multimodal retrieval-augmented generation (RAG) pipeline through a two-step training strategy combining supervised learning and reinforcement learning (RL), enhanced with Chain-of-Thought (CoT) reasoning for improved medical question answering.

\subsection*{Two-Step Training Strategy}

\begin{itemize}
    \item \textbf{Step 1: Supervised Pretraining.}  
    The model is initially trained using annotated medical data in multiple field with cross-entropy loss. During this stage, only the multilayer perceptron projector connecting Vision Tower and LLM is tuned.
    
    \item \textbf{Step 2: Reinforcement Fine-Tuning (RFT).}  
    The model is then refined using pre-built RTRA-format data to stimulate performance, as we are building roll-out data like RL. This is similar to building thinking-based models. During this stage, MLP and LLM parameters are unfrozen for training.

    All training stages are using cross-entropy loss for model performance optimization.
    \begin{equation}
        \mathcal{L}_{\text{CE}} = - \sum_{t=1}^{T} \log p_\theta(y_t | y_{<t}, q, \mathcal{D}),
    \end{equation}
    where $y_t$ is the target token, $q$ is the query, and $\mathcal{D}$ is the retrieved multimodal context.
    
\end{itemize}

\subsection*{Multimodal Integration and Vision Tower Tuning}

We follow the LLaVA-style autoregressive modeling to predict tokens given visual and textual inputs:
\begin{equation}
    \mathcal{L} = - \frac{1}{L} \sum_{t=1}^{L} \log P(x_t | x_{<t}, E_n; \theta),
\end{equation}
where $E_n$ are visual embeddings from both original and retrieved images.

To enhance task adaptation, we unfreeze the vision tower during instruction tuning. We test two variants:

\begin{itemize}
    \item \textbf{CLIP-based Encoders:}
    \begin{equation}
        L = \frac{1}{2N} \left( \sum_{i=1}^{N} - \log \frac{e^{s_{i,i}/\tau}}{\sum_{j=1}^{N} e^{s_{i,j}/\tau}} + \sum_{i=1}^{N} - \log \frac{e^{s_{i,i}/\tau}}{\sum_{j=1}^{N} e^{s_{j,i}/\tau}} \right),
    \end{equation}
    where $s_{i,j}$ is the similarity between image and text embeddings, and $\tau$ is a temperature parameter.

    \item \textbf{SigLip Encoders:}
    \begin{equation}
        L = -\frac{1}{|\mathbf{B}|} \sum_{i=1}^{\mathbf{B}} \sum_{j=1}^{\mathbf{B}} \log \left( \frac{1}{1 + \exp(z_{ij} (-t \mathbf{x_i} \cdot \mathbf{y_j} + b))} \right),
    \end{equation}
    where $z_{ij}$ indicates whether $(\mathbf{x_i}, \mathbf{y_j})$ is a positive match, $b$ is a learnable bias, and $t$ is a scaling factor.
\end{itemize}

Unfreezing the vision tower improves multimodal alignment, enhancing the model’s reasoning capabilities. In the next section, we evaluate performance against baselines and analyze the impact of each component.

\section{Experiments}
\label{sec:exp}

\begin{table*}[t]
\centering
    \resizebox{0.9\textwidth}{!}{
    \setlength{\tabcolsep}{0.2em}
    \begin{tabular}{c c c c c c c c c c}
        \toprule
         framework&    & BLEU1 & BLEU2 & BLEU3 & BLEU4 & METEOR &ROUGE\_L & CIDEr & SPICE \\
         \midrule
         LLaVA-1.5 Zeroshot\cite{liu2023improvedllava}& & 0.130 & 0.070 & 0.054 & 0.041 & 0.021 & 0.0821 & 0.053 & 0.046\\
         \midrule
         BiomedGPT\cite{zhang2023biomedgpt}& & 0.371	&0.232	&0.137	&0.073	&0.193	&0.231&	0.216	&0.184\\
         LLaVA-Med-7B\cite{li2023llavamed}& & 0.482&	0.337&	0.164&	0.108	&0.227	&0.273	&0.259	&0.267\\
         Huatuo-Vision-7B\cite{pubmedvision}& & 0.555&	0.414&	0.230&	0.155	&0.325	&0.291	&0.265	&0.305\\
         \midrule
         \textbf{MIRA (ours)}& & \textbf{0.571}	&\textbf{0.420}	&\textbf{0.259}	&\textbf{0.175}	&\textbf{0.292}&		\textbf{0.323}	&\textbf{0.299}	&\textbf{0.318}\\
         \bottomrule
    \end{tabular}
}
    \caption{Performance comparison of multimodal report generation abilities between MIRA and frameworks specialized on , measured on 1000 samples split from MIMIC-CXR .}
    \label{mimic_res}
    \vspace{-1.5em}
\end{table*}

\subsection{Datasets}

We evaluated our framework on these publicly available benchmark datasets.

\textbf{MIMIC-CXR.}
MIMIC-CXR~\cite{mimiccxr1} is a large-scale public dataset containing over 370,000 chest X-rays and corresponding radiology reports from 65,000+ patients. Released by MIT, it is widely used for tasks like disease classification, abnormality detection, and report generation, making it a key resource for multi-modal and vision-language research in medical imaging.

\textbf{PubMedVision.}
PubMedVision~\cite{pubmedvision} is a multi-modal benchmark built from the PMC-OA dataset, linking medical figures to their captions, abstracts, and full texts. It supports vision-language research, including pretraining and medical VQA, with diverse and well-aligned data across biomedical fields.

\subsection{Implementation Details}

We implement our frameworks using PyTorch and train them on NVIDIA A100 GPUs. For all experiments, we use the Adam optimizer with a learning rate of 1e-4 and a linear learning rate warmup over the first 10\% of training steps, followed by cosine decay. Input images are resized to 224×224 and normalized using ImageNet statistics, while textual inputs are tokenized using a domain-specific tokenizer aligned with the biomedical vocabulary. We employ mixed-precision training to accelerate computation and reduce memory consumption. For multi-modal pretraining, we adopt a batch size of 256 and train for 100 epochs, using early stopping based on validation loss. During fine-tuning, hyperparameters are optimized using grid search across different downstream tasks. All experiments are conducted with three random seeds to ensure reproducibility, and results are reported as the mean and standard deviation.

\subsection{Evaulation Metrics}

To rigorously evaluate the capabilities of MIRA across diverse biomedical tasks, we rely on task-specific metrics that capture both textual and visual understanding. For the report generation task on MIMIC-CXR, we employ a comprehensive set of natural language generation metrics, including BLEU, METEOR, ROUGE, CIDEr, and SPICE, each offering a unique perspective on fluency, relevance, and factual consistency between generated and reference reports. These metrics are standard in medical image captioning and allow us to benchmark performance against existing models with precision. In the context of visual question answering (VQA) on the PMC-VQA dataset, we use exact match accuracy as the principal metric. This measures whether the predicted answer matches the ground truth exactly, reflecting the model's ability to comprehend both visual content and medical semantics. Together, these evaluation metrics provide a robust framework to analyze the strengths and limitations of the proposed approach in real-world medical AI applications.


\subsection{Results on MIMIC-CXR}

We focus on evaluating the performance of the proposed MIRA framework on the MIMIC-CXR dataset, ensuring that the results are not biased by prior exposure. To achieve this, we avoid using general-purpose frameworks that might have already incorporated the MIMIC-CXR dataset during their pre-training. This preclusion is important as it ensures that the performance of the frameworks is evaluated on their ability to generalize to unseen data rather than benefiting from prior knowledge of the dataset. Instead, we select frameworks that are explicitly trained without the inclusion of MIMIC-CXR, thus offering a more genuine measure of their capability to handle medical image data and generate coherent reports based on their architectural design and pre-training strategies.
Table~\ref{mimic_res} compares MIRA with leading multimodal report generation frameworks on a 1,000-sample subset of MIMIC-CXR using standard metrics: BLEU (1–4), METEOR, ROUGE-L, CIDEr, and SPICE. These metrics evaluate lexical precision and semantic coherence—key factors for high-quality medical reports.

MIRA outperforms all baselines, achieving the highest BLEU-1 (0.571), BLEU-4 (0.175), METEOR (0.292), ROUGE-L (0.3232), CIDEr (0.299), and SPICE (0.318) scores. Its strong performance, despite no exposure to MIMIC-CXR during pre-training, reflects robust generalization rather than overfitting.
Compared to models like Huatuo-Vision-7B and LLaVA-Med-7B, MIRA shows notable gains, especially in CIDEr and SPICE—metrics crucial for generating informative and clinically relevant reports. Its performance is driven by its architecture, which excels at integrating visual and textual cues.
Frameworks like MMed-RAG were excluded due to unavailable code and pre-trained weights, limiting reproducibility. Unlike retrieval-heavy methods, MIRA offers a focused, transparent multimodal reasoning approach suited for medical image understanding.

\subsection{Results on PubMedVision}

Table~\ref{pmcvqares} summarizes MIRA’s performance on the PMC-VQA dataset, highlighting its strong multimodal reasoning in medical VQA tasks. MIRA achieves a correctness score of 0.66 (861/1297) on true/false questions, closely trailing top models like Qwen2.5-VL-72B and OpenAI o1 (both at 0.69), while outperforming LLaVA-Med (0.54) and BiomedGPT (0.58).
On detailed questions, MIRA scores 0.39 (385/971), ahead of Huatuo-Vision-7B (0.24) and XrayGPT (0.23), indicating its capacity for fine-grained visual understanding. However, it trails Qwen2.5-VL-72B (0.47) and OpenAI o1 (0.49), likely due to their significantly larger model sizes and computational resources.

\begin{table}[htbp]
    \centering
    \resizebox{\columnwidth}{!}{
    \begin{tabular}{c c c c}
    \toprule
         framework&    & Conv (1297 in total) & Details (971 in total, $c_s>0.5$) \\
         \midrule
         LLaVA-1.5 Zeroshot\cite{liu2023improvedllava}& & 0.45(578) & 0.02(20)\\
         \midrule
         BiomedGPT\cite{zhang2023biomedgpt}& & 0.58(752)	&0.21(201)\\
         LLaVA-Med-7B\cite{li2023llavamed}& & 0.54(700)& 0.19(183)	\\
         Huatuo-Vision-7B\cite{pubmedvision}& & 0.63(823)&	0.24(231)\\
         XrayGPT\cite{xraygpt}& & 0.59(769) &	0.23(222)\\
         \midrule
         Qwen2.5-VL-72B& & 0.69(892)&	0.47(457)\\
         OpenAI o1& & 0.69(893)&	0.49(480)\\
         Kimi-K1.5\cite{team2025kimi}& & 0.59(767)&	0.35(340)\\
         \midrule
       \textbf{MIRA (ours)}& & 0.66(861)	&0.39(385)	\\
       \bottomrule
    \end{tabular}
}
    \caption{Performance comparison of PMC-VQA question set correctness analysis}
    \label{pmcvqares}
    \vspace{- 1.5em}
\end{table}

MIRA balances efficiency and effectiveness, performing well in medical VQA without the massive scale of models like Qwen2.5-VL-72B. Its strong results—despite no pretraining on PMC-VQA—indicate genuine generalization. Fine-tuning on edge cases or integrating medical knowledge bases could further narrow the performance gap.

\subsection{Ablation Study}

\noindent\textbf{Preference Analysis.}
To further assess the contribution of specific components within MIRA-8B, an ablation study was conducted based on preference analysis against OpenAI o1 and Kimi-K1.5 frameworks, as shown in Table \ref{pref}. The results reveal that MIRA is preferred over OpenAI o1 in 46\% of cases (92 out of 200), suggesting that its design offers a balanced performance for PubMedVision QA pairs. However, when compared to Kimi-K1.5, MIRA-8B is favored more frequently, with a preference ratio of 59.5\% (119 out of 200), indicating that the framework's efficiency and multimodal reasoning capabilities provide a substantial advantage in this context. This ablation highlights the critical components of MIRA-8B that contribute to its generalization ability and its effectiveness across diverse PubMedVision tasks.

\begin{table}[htbp]
    \centering
    \resizebox{0.8\linewidth}{!}{
    \begin{tabular}{c c c c c}
         Comparison&   & OpenAI o1 & Kimi-K1.5\cite{team2025kimi}\\
         \midrule
         MIRA (ours)& & 0.460(92)	&0.595(119)	\\
    \end{tabular}
}
    \caption{Preference analysis on 200 sampled PubMedVision QA pairs shows that MIRA is favored by the judge model (Qwen2.5-VL-72B) in most cases.}
    \label{pref}
    \vspace{- 2em}
\end{table}
\begin{table}[b]
    \centering
    \resizebox{0.95\linewidth}{!}{
    \begin{tabular}{lcc}
        \toprule
        \textbf{Component} & \textbf{Conv (1297 in total)} & \textbf{Details (971 in total, $c_s>0.5$)} \\
        \midrule
        No Online (Local only)   & 0.49(639) & 0.11(108) \\
        No Offline (Search only)   & 0.55(714) & 0.23(222) \\
        Text Only                & 0.46(601) & 0.19(181) \\
        Vision Only              & 0.50(654) & 0.20(190) \\
        \bottomrule
    \end{tabular}
    }
    \caption{Ablation study showing performance under various RAG settings. “Conv” and “Details” represent the number of correct/favorited responses, which is the same setting as previous experiment in PMC-VQA.}
    \label{tab:rag_ablation}
\end{table}
\noindent\textbf{Attention Visualization.}

\begin{figure}[t]
    \centering
    \includegraphics[width=\linewidth]{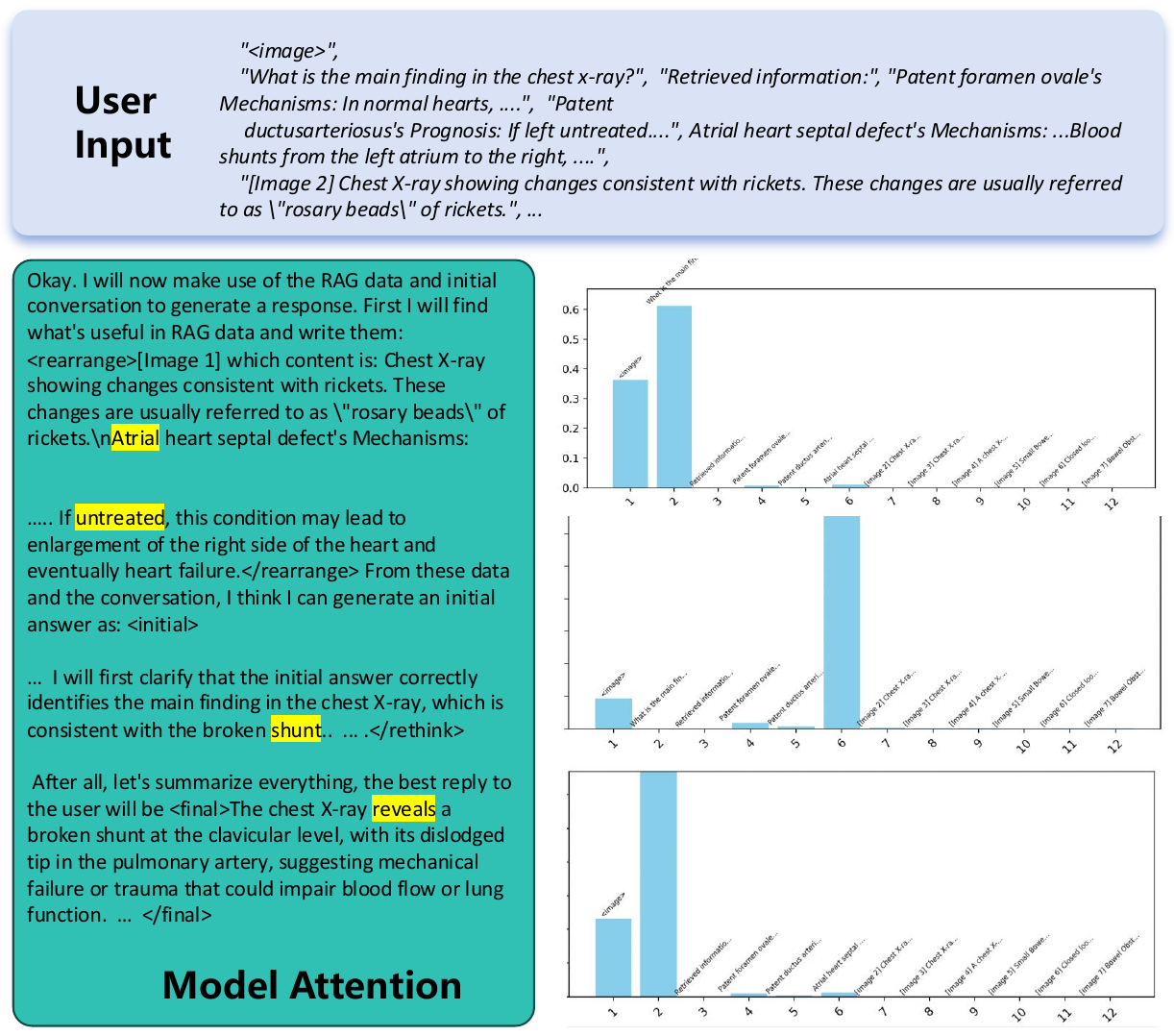}
    \caption{Visualization of attention distribution across all slices of input sequences, emphasizing the model's focus on critical tokens. This highlights how the model has learned selectively attends to important parts of the input, guiding the generation process for more accurate and contextually relevant responses.}
    \label{attention}
\end{figure}

To better understand how the framework utilizes retrieved information during response generation, we visualize the attention distribution over the input sources while decoding specific output tokens (highlighted in yellow) in Fig~\ref{attention}. The attention map shows when inferencing the tokens highlighted in yellow, the model's attention distribution in the user input's components, respectively, input iamges, query, and each slice of RAG result. The map only calculates attention in user input, while normalizing all these attention components to a sum of 1. Our analysis shows that the framework's attention is initially focused on the visual input, particularly around regions describing "chest X-ray showing changes consistent with rickets" and the anatomical location ("clavicular level"). As the framework proceeds to generate its response, attention shifts toward textual cues in the retrieved documents—especially phrases such as "atrial septal defect’s mechanisms", "if untreated", and "shunt". These elements are semantically aligned with the generated conclusion involving a broken shunt and its clinical implications. Note that the framework predominantly focuses its attention on the response-side context during decoding, leading to an uneven distribution across inputs. For this visualization, we normalize the attention weights across source tokens to highlight their relative importance in guiding generation, based on training-time behavior. Brief spikes on unrelated regions suggest room for improving alignment for better coherence and factual accuracy in complex medical cases.

\noindent\textbf{Component Analysis}
To understand the contribution of different RAG components, we perform an ablation study under various settings: disabling online retrieval (local only), disabling offline DDGS memory, and isolating modalities (text-only and vision-only). As shown in Table~\ref{tab:rag_ablation}, removing the online component leads to a noticeable drop in performance, particularly in detail-level understanding, indicating the importance of up-to-date retrieval. Similarly, the absence of offline DDGS memory reduces overall grounding and specificity. Interestingly, vision-only settings outperform text-only in both conversational and detailed responses, highlighting the framework’s strength in visual reasoning. However, the best performance is achieved when both retrieval pathways and modalities are integrated.

\section{Conclusion}
\label{sec:conclusion}

We present MIRA, a novel approach to improving the factuality and reliability of Multimodal Large Language Models (Multimodal-LLMs) in medical Retrieval-Augmented Generation (RAG). MIRA addresses two key challenges: mitigating factuality risks by rearranging retrieved contexts for relevance and reliability, and reducing excessive dependence on external data to prevent overfitting. By balancing intrinsic medical knowledge with external insights, MIRA generates more accurate and contextually grounded medical responses using our RtRa framework. Our experiments and ablation studies demonstrate significant improvements in factuality, clarity, and robustness, paving the way for more reliable AI systems in healthcare settings. Future work may focus on model fine-tuning and integration with specialized medical knowledge bases to increase performance in even more complex medical scenarios.

\bibliographystyle{ACM-Reference-Format}
\bibliography{sample-base}

\appendix

\onecolumn
\subsection*{Appendix}
\begin{table*}[ht]
\centering
\caption{Hyperparameter Settings for LLaVA Training}
\label{tab:hyperparams}
\begin{tabular}{cc}
\toprule
\textbf{Hyperparameter} & \textbf{Value} \\
\midrule
\texttt{version(llava\_prompt\_version)} & \texttt{RTRA-Format-v1} \\
\texttt{mm\_tunable\_parts} & \texttt{mm\_mlp\_adapter} in Pretraining, \texttt{mm\_mlp\_adapter, mm\_language\_model} in RFT \\
\texttt{deepspeed\_mode} & \texttt{zero\_2\_offload} in both pretraining and finetuning \\
\texttt{mm\_vision\_tower\_lr} & \texttt{2e-6} \\
\texttt{vision\_tower} & \texttt{google/siglip-so400m-patch14-384} \\
\texttt{mm\_projector\_type} & \texttt{mlp2x\_gelu} \\
\texttt{mm\_vision\_select\_layer} & \texttt{-2} \\
\texttt{rag\_enabled} & \texttt{True} \\
\texttt{rag\_tokenizer} & \texttt{all-MiniLM-L6-v2} \\
\texttt{rag\_topk} & \texttt{3} \\
\texttt{query\_rewrite\_enabled} & \texttt{True} \\
\texttt{query\_rewrite\_model} & \texttt{Qwen-2-5-8B} \\
\texttt{image\_aspect\_ratio} & \texttt{anyres} \\
\texttt{image\_grid\_pinpoints} & \texttt{[(384, 768), (768, 384), (768, 768), (1152, 384), (384, 1152)]} \\
\texttt{mm\_patch\_merge\_type} & \texttt{spatial\_unpad} \\
\texttt{precision} & \texttt{bfloat16} \\
\texttt{Pretrain\_num\_train\_epochs} & \texttt{1} \\
\texttt{Pretrain\_per\_device\_train\_batch\_size} & \texttt{4} \\
\texttt{Pretrain\_per\_device\_eval\_batch\_size} & \texttt{1} \\
\texttt{Pretrain\_gradient\_accumulation\_steps} & \texttt{16} \\
\texttt{RFT\_num\_train\_epochs} & \texttt{2} \\
\texttt{RFT\_per\_device\_train\_batch\_size} & \texttt{1} \\
\texttt{RFT\_per\_device\_eval\_batch\_size} & \texttt{1} \\
\texttt{RFT\_gradient\_accumulation\_steps} & \texttt{32} \\
\texttt{learning\_rate} & \texttt{1e-5 in pretraining, 2e-6 in finetuning} \\
\texttt{weight\_decay} & \texttt{0.0} \\
\texttt{warmup\_ratio} & \texttt{0.03} \\
\texttt{lr\_scheduler\_type} & \texttt{cosine} \\
\texttt{tf32} & \texttt{True} \\
\texttt{model\_max\_length} & \texttt{32768} \\
\texttt{torch\_compile\_backend} & \texttt{inductor} \\
\texttt{dataloader\_drop\_last} & \texttt{True} \\
\texttt{attn\_implementation} & \texttt{sdpa} \\
\bottomrule

\end{tabular}
\end{table*}

\subsubsection*{Notes}
\begin{itemize}
    \item The experiments were conducted using 4$\times$A100 80GB GPUs in RFT stage, and 4$\times$ A100 40GB GPUs in Pretraining stage.
    \item For query rewrite, zero-shot models could be used for summarizing user inputs. \texttt{Qwen-2-5-8B} is recommended, and \texttt{Mistral-small-22B} is also tried during development. The external model is hosted using \texttt{Ollama}.
\end{itemize}

\newpage
\subsection*{Additional Dataset Visualizations}
\begin{convbox}
    \begin{center}
        \color{headerblue}
        \LARGE \textbf{RTRA DataSlice Example 1} \\
    \end{center}
    \vspace{0.2cm}

    \small
    \textbf{\color{headerblue} Human Query:} \\
    Is it a malignant lesion? \\
    \vspace{0.1cm}
    \textbf{\color{headerblue}MIRA Response:} \\
    \begin{itemize}[leftmargin=1cm, label={\faAngleRight}]
        \item \textit{Rearrange:} Selected:
            \begin{enumerate}[leftmargin=*]
                \item Cryptococcus neoformans and C. gattii often cause invasive lung infections in adults with HIV or other weakened immune systems. It can look like lung cancer or tuberculosis on scans.
                \item <Image 4> CT scan showing cryptogenic organizing pneumonia (biopsy-proven).
            \end{enumerate}
        \item \textit{Initial:} The lesion identified in the CT scan, initially diagnosed as cryptogenic organizing pneumonia, is not definitively a malignant lesion. Cryptogenic organizing pneumonia is a type of lung inflammation that can mimic the appearance of cancer on imaging studies, but it is not a form of cancer. However, to rule out malignancy, additional tests such as a biopsy or further imaging may be necessary.
        \item \textit{Rethink:} 
              \begin{enumerate}[leftmargin=*]
                  \item I'll confirm the initial diagnosis and the nature of the lesion, emphasizing that cryptogenic organizing pneumonia (COP) is a condition that can mimic the appearance of a malignant lesion on imaging but is not cancerous.
                  \item I'll clarify that while COP is the initial diagnosis, the possibility of a malignant lesion cannot be entirely ruled out without further diagnostic tests, such as a biopsy or additional imaging studies.
                  \item I'll mention that the differential diagnosis should also consider other conditions that can present similarly, such as pulmonary cryptococcosis, especially in immunocompromised individuals, and highlight the importance of a comprehensive evaluation.
              \end{enumerate}
        \item \textit{Final:} The lesion was initially diagnosed as cryptogenic organizing pneumonia, a non-malignant lung condition. However, a definitive diagnosis would require further evaluation and tests, such as a biopsy, to confirm the nature of the lesion and rule out any malignancy. Additionally, conditions like pulmonary cryptococcosis, which can also mimic cancer on imaging, should be considered, especially in immunocompromised individuals. It's always best to consult with a healthcare professional for a thorough assessment and accurate diagnosis.   \end{itemize}
    \vspace{0.15cm}
    \textbf{\color{headerblue} RAG Sequence For the Query:} \\
    \begin{itemize}[leftmargin=1cm, label={\faDatabase}]
        \item Nonspecific interstitial pneumonia's Diagnosis:  Diagnosis is made by reviewing patient history, imaging, lung function tests, and sometimes a surgical lung biopsy. Lung biopsies show two patterns: cellular (with inflammation but minimal fibrosis) and fibrosing (with interstitial fibrosis and varying levels of inflammation). Both patterns are uniform and lack the prominent fibroblastic foci found in other types of idiopathic interstitial pneumonia.
        \item Cryptococcosis's Pulmonary cryptococcosis:  Cryptococcus neoformans and C. gattii often cause invasive lung infections in adults with HIV or other weakened immune systems. While healthy adults may get it too, they usually have mild symptoms and don't need treatment. However, these infections can spread to the brain, especially in immunocompromised individuals. Pulmonary cryptococcosis is common worldwide but often underdiagnosed due to limitations in testing. It can look like lung cancer or tuberculosis on scans. The CrAg test and blood cultures are usually negative unless the infection has spread. If not treated, pulmonary cryptococcosis can make cryptococcal meningitis worse.
        \item C: Cryptococcal meningitis's Pulmonary cryptococcosis:  Cryptococcus (both C. neoformans and C. gattii) commonly causes lung infections in adults with HIV or other weakened immune systems, but it can also affect healthy adults with milder symptoms. While these infections may not need treatment, careful monitoring is important. In some cases, the infection can spread to the brain and nervous system, especially in those with weakened immune systems. Pulmonary cryptococcosis has a worldwide presence but is often underdiagnosed due to diagnostic limitations. It typically presents as lung nodules that can mimic cancer, tuberculosis (TB), or other fungal infections. However, tests like blood cultures and the Cryptococcal antigen test are usually negative unless the infection has spread. Additionally, if not managed properly, this condition can worsen the prognosis of cryptococcal meningitis.
        \item Online RAG: hamburger line 1 hamburger line 2 hamburger line 3 Close Home (https://www.belmarrahealth.com/) » Brain Function (https://www.belmarrahealth.com/brain-function/) » Encephalomalacia: Definition, causes, types, symptoms, and treatment Encephalomalacia is a type of serious brain damage that results in the softening or loss of brain tissue. Causes of encephalomalacia are often linked to inflammation or hemorrhages that are a consequence of being afflicted by cerebral infarction, cerebral ischemia (https://www.belmarrahealth.com/types-cerebral-ischemia-symptoms-prevention-treatment-tips/) , infection, craniocerebral trauma, or another injury. The term, encephalomalacia, is often used upon gross pathological inspection, as it is much easier to better describe blurred cortical margins and decreased consistency of brain tissue, or brain softening. Advertisement [cky\_video\_placeholder\_title] Softening may be seen in a localized area or be widespread, depending on the underlying cause. It can affect virtually any part of the brain, possibly damaging tissues of the frontal, occipital, parietal, and temporal lobe. This can present with several motor and sensory deficits. Considered the most common cause of encephalomalacia, strokes occur due to the interrupted blood supply to the brain either through a lack of blood supply to brain tissue or due to hemorrhage of a brain blood vessel. Oxygenated blood is vital for the function and maintenance of brain cells, which will become damaged or die if the blood supply is not promptly restored. Occurs when the flow of blood in the brain has become interrupted. This may be due to abnormal brain swelling or the removal of brain tumors resulting in brain damage. Damage to brain tissue due to conditions such as a stroke can result in the formation of scar tissue. This area of tissue undergoes contractions and eventually forms encephalomalacia within the brain Blunt force trauma to the skull will lead to contusions and possibly brain damage if the force is strong enough. Other forms of penetrating trauma, like from a knife or bullet to the skull, may also lead to the development of encephalomalacia Encephalomalacia may also be the result of: Also known as cerebrocortical necrosis (CCN), this neurological disease is caused by the disruption of thiamine production. Thiamine is a B vitamin (B1) and a key chemical in glucose metabolism. If a person were to become deficient in thiamine, significant side effects are likely to develop. The most threatening of which is the inhibition of neurological activity. Polioencephalomalacia will result in damage to the gray matter of the central nervous system, where the majority of information processing.
    \end{itemize}
\end{convbox}

\begin{convbox}
    \begin{itemize}[leftmargin=*]
        \item Image 2: CT of the chest demonstrating right-sided pneumonia (left side of the image).
        \item Image 3: Cavitating pneumonia due to MRSA as seen on a CT scan.
        \item Image 4: CT scan showing cryptogenic organizing pneumonia (biopsy-proven).
        \item Image 5: MRI of Brain and MRA Brain showing encephalomalacia with extensive changes.
        \item Image 6: Not retrieved.
        \item Image 7: Not retrieved.
    \end{itemize}
\end{convbox}

\begin{figure}[h]
    \centering
    \begin{minipage}{0.48\textwidth}
        \centering
        \includegraphics[width=\textwidth,height=4.5cm]{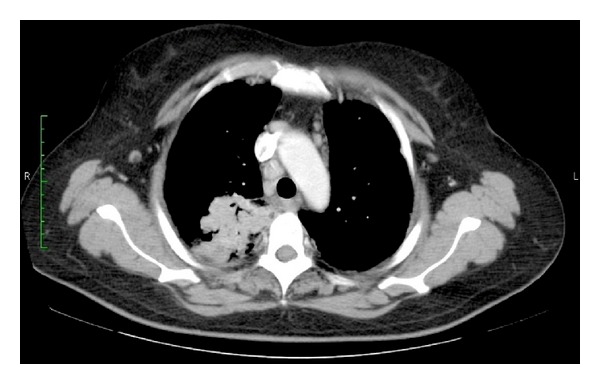}
        \caption*{Input Image (User-Provided)}
    \end{minipage}\hfill
    \begin{minipage}{0.48\textwidth}
        \centering
        \begin{tabular}{cc}
            \includegraphics[width=0.22\textwidth,height=1.8cm]{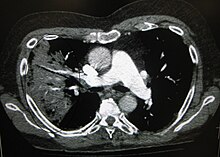} &
            \includegraphics[width=0.22\textwidth,height=1.8cm]{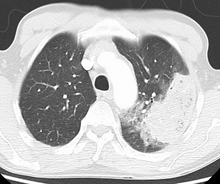} \\
            \scriptsize RAG Image 2 & \scriptsize RAG Image 3 \\
            \includegraphics[width=0.22\textwidth,height=1.8cm]{ACM_MM/rtra_data_visualization/source_7/b.jpg} &
            \includegraphics[width=0.22\textwidth,height=1.8cm]{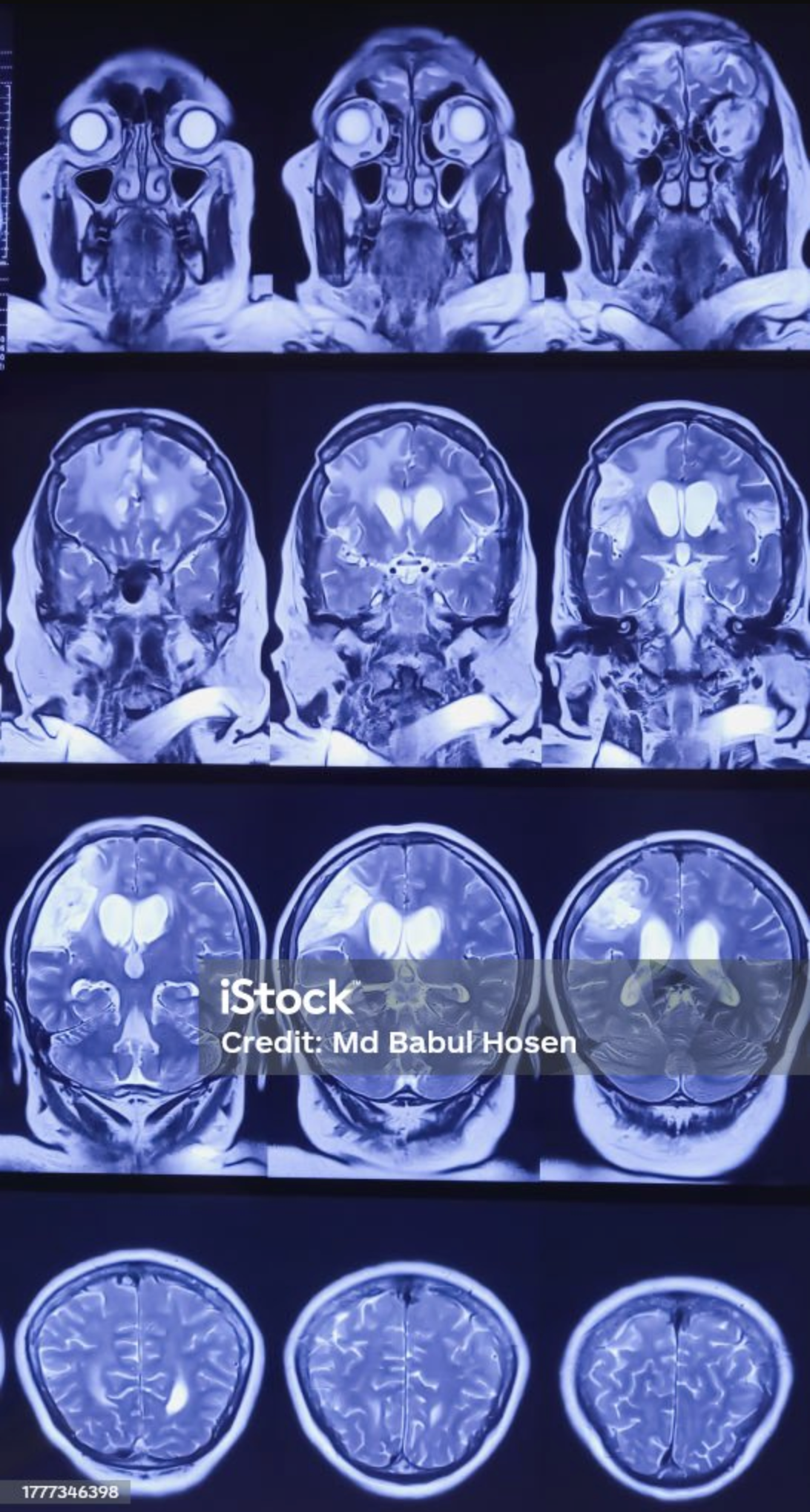} \\
            \scriptsize RAG Image 4 & \scriptsize Online RAGImg 5\\
        \end{tabular}
    \end{minipage}
\end{figure}

\begin{convbox}
    \begin{center}
        \color{headerblue}
        \LARGE \textbf{RTRA DataSlice Example 2} \\
    \end{center}
    \vspace{0.2cm}

    \small
    \textbf{\color{headerblue} Human Query:} \\
    What is the image showing? \\
    \vspace{0.1cm}
    \textbf{\color{headerblue}MIRA Response:} \\
    \begin{itemize}[leftmargin=1cm, label={\faAngleRight}]
        \item \textit{Rearrange:} Selected:
            \begin{enumerate}[leftmargin=*]
                \item Brachydactyly type D:  Brachydactyly type D, also known as short thumb or clubbed thumb, is a genetic trait where the thumb appears shorter and wider with a broader nail bed than usual. The distal phalanx of the affected thumb is approximately two-thirds the length of a normal thumb. It is quite common, affecting about 2–3\% of the population, and is associated with the HOXD13 gene on chromosome 2q31.1.
                \item <Image 2> Merkel cell carcinoma. Gross pathology specimen.
            \end{enumerate}
        \item \textit{Initial:} The image is showing a gross pathology specimen of Merkel cell carcinoma.
        \item \textit{Rethink:} 
              \begin{enumerate}[leftmargin=*]
                  \item I'll confirm the accuracy of the initial answer by verifying that the image indeed shows a gross pathology specimen of Merkel cell carcinoma.
                  \item I'll enhance the initial answer by providing a brief description of what Merkel cell carcinoma is, including its clinical significance and common presentation.
                  \item I'll ensure the answer is concise and directly addresses the question, while also offering enough context for the user to understand the significance of the image.
              \end{enumerate}
        \item \textit{Final:} The image is showing a gross pathology specimen of a mass, specifically Merkel cell carcinoma, which has been resected. Merkel cell carcinoma is a rare and aggressive type of skin cancer that typically appears as a firm, flesh-colored or bluish-red nodule on sun-exposed areas. Early detection and treatment are crucial due to its rapid growth and potential to spread to other parts of the body.
    \end{itemize}
\end{convbox}

\begin{convbox}
    \vspace{0.15cm}
    \textbf{\color{headerblue} RAG Sequence For the Query:} \\
    \begin{itemize}[leftmargin=1cm, label={\faDatabase}]
        \item Carpal tunnel syndrome's Anatomy:  The carpal tunnel is a small space in the wrist, just above the base of the palm. It contains nine flexor tendons (which bend the fingers) and the median nerve. This nerve provides feeling to the thumb, index, middle, and half of the ring finger, as well as controls muscles that move the thumb. The carpal tunnel is bordered by bones and a strong ligament called the transverse carpal ligament or flexor retinaculum. It is located in the middle third of the palm, between the scaphoid bone (at the base of the thumb) and the hamate hook (along the ring finger).
        \item Brachydactyly type D:  Brachydactyly type D, also known as short thumb or clubbed thumb, is a genetic trait where the thumb appears shorter and wider with a broader nail bed than usual. The distal phalanx of the affected thumb is approximately two-thirds the length of a normal thumb. It is quite common, affecting about 2–3
        \item C: Brachydactyly type D's Signs and symptoms:  Brachydactyly type D is a skeletal condition where the tip of the thumb appears shorter due to partial fusion or early closure of the growth plate in the distal phalanx. This can affect one or both thumbs, making them approximately half to two-thirds the length of normal thumbs.
        \item Online RAG: The gross pathology specimen of Merkel cell carcinoma presents as a firm, nodular mass with a tan to pink cut surface. It is typically well-circumscribed but may show areas of ill-defined infiltration into surrounding tissue. On sectioning, the mass appears solid and homogeneous, sometimes with regions of hemorrhage or necrosis, especially in larger tumors. The surface may be smooth or lobulated, and involvement of the dermis and subcutaneous tissue is common. Despite its potentially benign appearance on gross examination, Merkel cell carcinoma is a highly aggressive neuroendocrine skin tumor often arising in sun-exposed areas of elderly or immunocompromised individuals.
        \item Image 2: Merkel cell carcinoma. Gross pathology specimen.
        \item Image 3: The cut surface of desmoid-type fibromatosis is firm, white, and whorled. The white tumor infiltrates the adjacent skeletal muscle (red tissue – lower left) and fat (yellow tissue – upper left). This tendency for invasion of adjacent normal tissues and structures is the reason that desmoid-type fibromatosis has a relatively high rate of local recurrence, even after surgical removal.
        \item Image 4: Image showing Gastrointestinal Stromal Tumor after surgical removal.
        \item Image 5: Squamous-Cell Carcinoma of the Skin | NEJM.
        \item Image 6: Radiology Workflow | Efficiencies
        \item Image 7: Patient enrollment workflow. MRI, magnetic resonance imaging; SLNB ...
    \end{itemize}
\end{convbox}

\begin{figure}[h]
    \centering
    \begin{minipage}{0.48\textwidth}
        \centering
        \includegraphics[width=\textwidth,height=4.2cm]{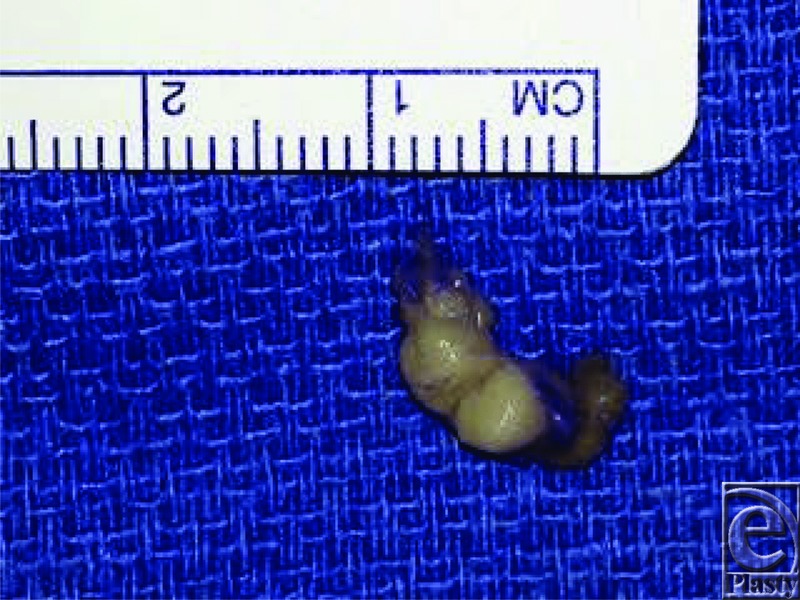}
        \caption*{Input Image (User-Provided)}
    \end{minipage}\hfill
    \begin{minipage}{0.48\textwidth}
        \centering
        \begin{tabular}{ccc}
            \includegraphics[width=0.31\textwidth,height=1.6cm]{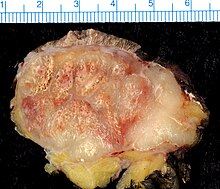} &
            \includegraphics[width=0.31\textwidth,height=1.6cm]{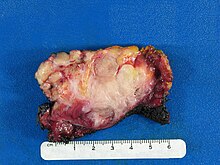} &
            \includegraphics[width=0.31\textwidth,height=1.6cm]{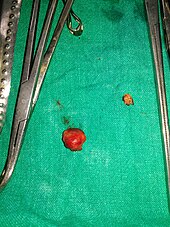} \\
            \scriptsize RAG Image 2 & \scriptsize RAG Image 3 & \scriptsize RAG Image 4 \\
            \includegraphics[width=0.31\textwidth,height=1.6cm]{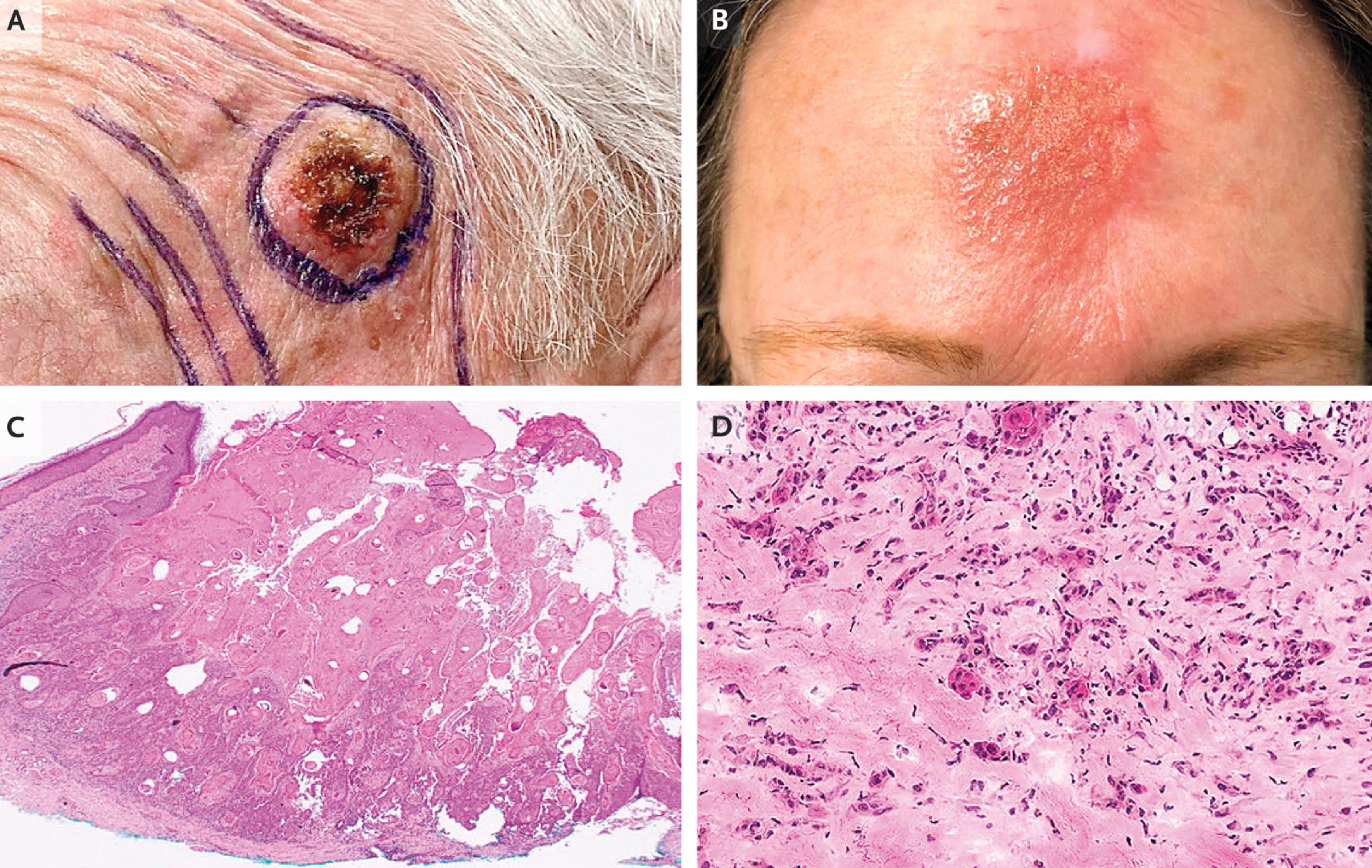} &
            \includegraphics[width=0.31\textwidth,height=1.6cm]{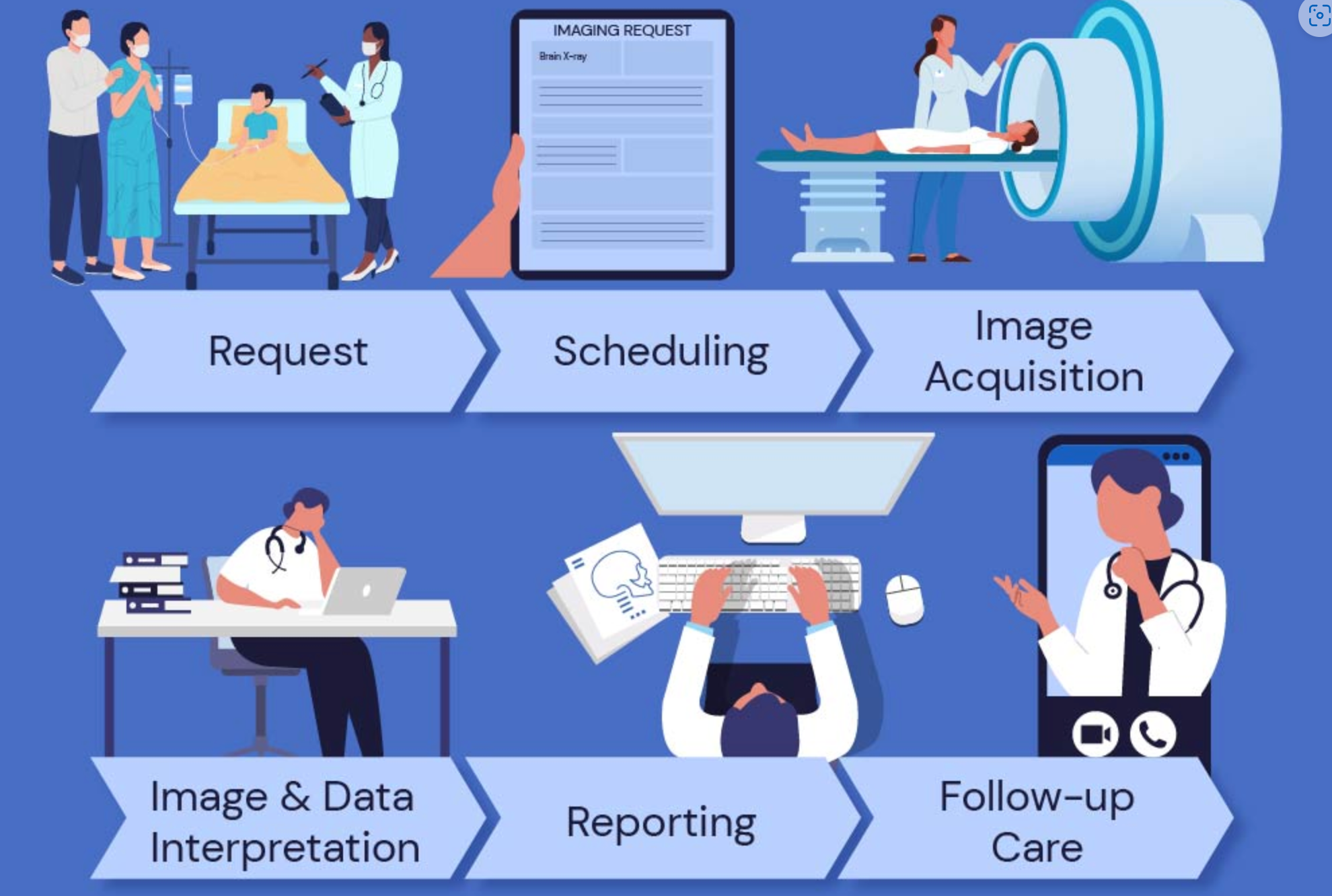} &
            \includegraphics[width=0.31\textwidth,height=1.6cm]{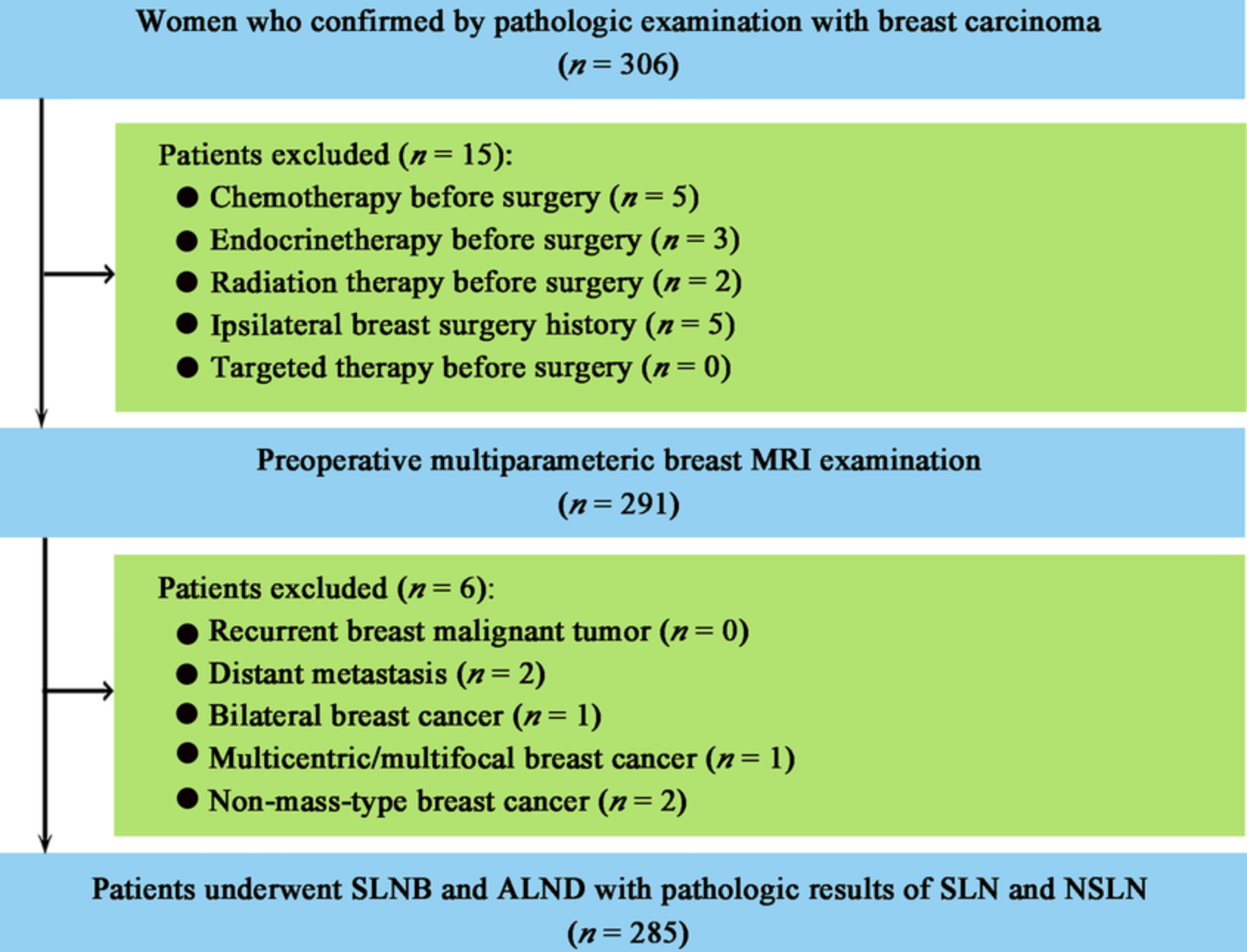} \\
            \scriptsize Online RAGImg 5 & \scriptsize Online RAGImg 6 & \scriptsize Online RAGImg 7 \\
        \end{tabular}
    \end{minipage}
\end{figure}

\end{document}